\documentclass[11pt]{article}

\usepackage[margin=1in]{geometry}
\usepackage{microtype}
\usepackage{graphicx}
\usepackage{booktabs}
\usepackage{hyperref}
\usepackage{xcolor}
\usepackage{float}
\usepackage{amsmath}
\usepackage{amssymb}
\usepackage[numbers]{natbib}

\usepackage{amsmath,amsfonts,bm}

\def\eqref#1{equation~\ref{#1}}

\def\1{\bm{1}}

\DeclareMathAlphabet{\mathsfit}{\encodingdefault}{\sfdefault}{m}{sl}
\SetMathAlphabet{\mathsfit}{bold}{\encodingdefault}{\sfdefault}{bx}{n}

\newcommand{\ie}{\textit{i}.\textit{e}., }
\newcommand{\eg}{\textit{e}.\textit{g}., }
\newcommand{\ours}{\textsc{CRTBench}}

\title{Controlled Reformulation Testing for Logical Consistency in Large Language Models}

\author{
Alexander Gu\\
University of Texas at Austin\\
\and
Alan Chen\\
University of Texas at Austin
}

\date{}

\begin{document}

\maketitle

\begin{abstract}
Large language models (LLMs) frequently contradict themselves when the surface form of a logically equivalent question changes. We present a benchmark of 350 question families (1,750 total questions) for Controlled Reformulation Testing (\ours{}) to evaluate logical invariance. In this benchmark, we investigate LLMs' ability to maintain consistent answers across controlled reformulations, which include contrapositive rewriting, double negation, negation flipping, and passive voice. We evaluate several frontier LLMs and observe an accuracy-consistency gap where GPT-5.4-mini achieves 98.9\% base accuracy but only 60.3\% family-level consistency, while reasoning-optimized o4-mini achieves 96.9\% consistency. From our experiments, we observe that failures cluster around logically nontrivial transformations such as contrapositive rewriting (72.4\% for GPT-5.4-mini) and double negation (84.6\%), while surface-level rephrasing remains robust (94--100\%). Increasing reasoning effort improves GPT-5.4-mini to 85.4\% consistency, but leaves GPT-5.4 unchanged overall because gains on nested negation are offset by failures on quantifier families. These results show that accuracy alone is not enough for evaluating logical reasoning in LLMs.
\end{abstract}

\section{Introduction}

High accuracy on logical reasoning benchmarks does not imply that a model answers logically equivalent questions consistently.
For reasoning systems, logical invariance is a fundamental requirement where logically equivalent questions should receive consistent answers regardless of surface phrasing.
A model that answers ``Is it true that cats are mammals?'' with ``Yes'' while answering its double negation ``Is it NOT the case that it is NOT true that cats are mammals?'' with ``No'' exhibits a logical contradiction that undermines its ability as a true reasoning system.

Prior efforts document LLM inconsistency through arbitrary paraphrases \citep{shi2025ppcv, jang2025selfconsistency} or faithfulness tests \citep{turpin2023faithfulness}, but these approaches cannot identify which logical operations cause failures. 
Unlike paraphrase-based evaluations, our transformations are logically grounded and isolate specific operations (\eg negation composition, contraposition), enabling causal attribution of failures. 
Existing logical reasoning benchmarks such as LogicBench \citep{parmar2024logicbench} and FOLIO \citep{han2022folio} evaluate whether LLMs reach correct answers, but do not test whether models remain consistent across logically equivalent reformulations of the same question, which is what we directly measure.
We introduce \ours{}, a benchmark consisting of controlled logical transformations such as contrapositive ($P{\to}Q \Leftrightarrow \neg Q{\to}\neg P$), double negation ($P \Leftrightarrow \neg\neg P$), De Morgan's laws, quantifier rewriting, negation flipping (answer-reversing), and passive voice (surface control) to analyze and classify invariance failures at the transformation level.

Our contributions are as follows: \textbf{(1)} We propose a benchmark of 350 question families spanning seven balanced reasoning categories with controlled logical reformulations. 
\textbf{(2)} We experiment on frontier LLMs with evidence of an accuracy-consistency gap. 
\textbf{(3)} We perform transformation- and category-level analysis showing that negation-heavy operations, rather than surface rephrasing, drive inconsistency, while quantifier reasoning emerges as a distinct failure mode for reasoning-enabled GPT-5.4.

\section{Benchmark Design}

A \emph{question family} consists of a base question and four reformulations, each applying a specific logical transformation. 
We define three transformation classes:

\begin{itemize}
    \item \textbf{Equivalence-preserving}: contrapositive, double negation, De Morgan, quantifier rewrite, passive voice. In this case, the expected answer matches the base.
    \item \textbf{Answer-reversing}: negation flip. Here, the expected answer is the logical complement (Yes$\leftrightarrow$No; ``Cannot be determined'' stays unchanged).
    \item \textbf{Surface control}: passive voice. Surface rephrasing preserves logical structure.
\end{itemize}

\paragraph{Metrics.} Our first metric, \emph{Base accuracy}, measures whether the model answers the original (base) question correctly. 
The second, \emph{Family consistency}, is stricter as it requires the base answer and all four reformulation answers to match their respective labels. 
A family is consistent only if the model gets every version right (\ie one wrong answer out of five would mean the family is inconsistent). 
This makes the gap between the two metrics directly interpretable.
For example, a model with 99\% accuracy but 60\% family consistency gets individual questions right but still contradicts itself across reformulations roughly two-fifths of the time.

We generate 350 families across seven categories: propositional logic (50), syllogistic reasoning (50), De Morgan's laws (50), quantifier reasoning (50), multi-hop inference (50), nested negation (50), and conditional fallacies (50), yielding 1,750 total questions. 
Not all transformations apply to every category; in the final benchmark, contrapositive applies to 290 families, passive voice to 300, De Morgan to 50, and quantifier rewriting to 60, while double negation and negation flip apply to all 350 families.

\paragraph{Validation protocol.} We validated the benchmark in two stages. First, we generated a base-only review set and manually reviewed all 350 base questions for logical correctness, unambiguous gold labels, and obvious grammatical awkwardness before any reformulations were added. Second, after generating the reformulations, we manually audited question families for transformation fidelity, answer-label correctness, and local linguistic naturalness. Equivalence-preserving transformations were checked to preserve truth conditions, while answer-reversing transformations were checked against explicitly mapped labels rather than a generic Yes/No inversion. Families with awkward phrasing or incorrect labels were revised before final evaluation. We did not collect formal inter-annotator agreement, so we treat this as internal quality control rather than a separate human-naturalness study.

\begin{table*}[h]
    \centering
    \caption{Model performance on \ours{} ($n$=350 families, zero-shot). Columns report base accuracy, family consistency, and per-transformation consistency. Best values are in \textbf{bold}. (reas.) = \texttt{reasoning\_effort=high}.}
    \label{tab:main}
    \footnotesize
    \setlength{\tabcolsep}{3pt}
    \resizebox{\textwidth}{!}{%
    
    \begin{tabular}{lcccccccc@{}}
    \toprule
    \textbf{Model} & \textbf{{Base Acc.}} & \textbf{Family Cons.} & \textbf{Contrap.} & \textbf{De Morgan} & \textbf{Double Neg.} & \textbf{Neg. Flip} & \textbf{Passive} & \textbf{Quant. Rewrite} \\
    \midrule
    GPT-5.4-mini & \textbf{98.9} & 60.3 & 72.4 & \textbf{100} & 84.6 & 94.6 & 98.7 & 90.0 \\
    GPT-5.4-mini (reas.) & 96.0 & 85.4 & 94.5 & \textbf{100} & 92.3 & 94.0 & 98.0 & 70.0 \\
    GPT-5.4 & \textbf{98.9} & 86.6 & 93.8 & \textbf{100} & 93.1 & 98.6 & 98.7 & 91.7 \\
    GPT-5.4 (reas.) & 92.9 & 86.6 & 93.4 & \textbf{100} & 95.4 & 92.3 & 99.0 & 78.3 \\
    Claude Sonnet 4 & 95.1 & 86.9 & 92.8 & 96.0 & 93.1 & 96.9 & 94.0 & \textbf{93.3} \\
    Gemini 2.5 Flash & 96.9 & 89.4 & 95.9 & \textbf{100} & 96.0 & 98.6 & 94.3 & 83.1 \\
    o4-mini & 98.6 & \textbf{96.9} & \textbf{98.6} & \textbf{100} & \textbf{98.0} & \textbf{99.1} & \textbf{99.7} & 90.0 \\
    \bottomrule
    \end{tabular}
    }
\end{table*}

\section{Experiments}

\subsection{Setup}
We evaluate frontier LLMs under zero-shot prompting scenarios (Temperature $T{=}0$). GPT-5.4 and GPT-5.4-mini at two reasoning levels (\texttt{reasoning\_effort=none} and \texttt{high}), Claude Sonnet 4 (Anthropic), Gemini 2.5 Flash (Google), and o4-mini (OpenAI, reasoning-optimized).
We also prompt models to directly answer ``Yes'', ``No'', or ``Cannot be determined'' and extract answers using regex. 
The regex-based extraction failure rate is effectively zero as all models except Gemini 2.5 Flash were fully parseable, and Gemini produced 5 \texttt{UNCLEAR} outputs total (0.29\% of questions).

\subsection{Main Results}    

Table~\ref{tab:main} presents the results across all 350 families\footnote{With $n{=}350$, a 95\% Confidence Interval (CI) for proportion $p$ is ${\approx}\,p \pm 1.96\sqrt{p(1{-}p)/350}$. \eg GPT-5.4-mini's 60.3\% has CI $\pm$5.1pp.}. 
From this, we emphasize three key findings:

\textbf{Accuracy $\neq$ consistency.} GPT-5.4-mini achieves 98.9\% base accuracy but only 60.3\% family consistency (38.6pp gap). GPT-5.4 shows 98.9\% accuracy but only 86.6\% consistency. All models exhibit lower consistency than base accuracy. Models answer individual questions correctly while contradicting themselves across reformulations.

\textbf{o4-mini remains strongest overall, but reasoning effort helps unevenly.} o4-mini achieves 96.9\% family consistency vs.\ 60.3\% for GPT-5.4-mini. Enabling reasoning on GPT-5.4-mini closes much of this gap (60.3$\to$85.4\%), but GPT-5.4 itself does not improve overall, remaining at 86.6\% family consistency.

\textbf{Failures are transformation-specific.} Contrapositive rewriting is hardest for GPT-5.4-mini (72.4\%), while passive voice remains 98.7\%. The 26.3pp gap between these indicates that logical, not surface, transformations are the source of inconsistency. Double negation also remains difficult for GPT-5.4-mini (84.6\%), while o4-mini stays near ceiling on all core transformations (98.0--99.7\%).


\begin{table*}[h]
\centering
\footnotesize
\caption{Per-transformation consistency (\%) on the four core transformations in the main text: contrapositive ($n{=}290$), double negation ($n{=}350$), negation flipping ($n{=}350$), and passive voice ($n{=}300$). Best in \textbf{bold}. (r)=reasoning\_effort=high.}
\label{tab:trans}
\setlength{\tabcolsep}{3pt}
\resizebox{\textwidth}{!}{%
\begin{tabular}{lccccccc@{}}
\toprule
\textbf{Transformation} & \textbf{GPT 5.4-mini} & \textbf{GPT 5.4-mini(r)} & \textbf{GPT 5.4} & \textbf{GPT 5.4(r)} & \textbf{Claude Sonnet 4} & \textbf{Gemini 2.5 Flash} & \textbf{o4-mini} \\
\midrule
Contrapositive & 72.4 & 94.5 & 93.8 & 93.4 & 92.8 & 95.9 & \textbf{98.6} \\
Double Negation & 84.6 & 92.3 & 93.1 & 95.4 & 93.1 & 96.0 & \textbf{98.0} \\
Negation Flipping & 94.6 & 94.0 & 98.6 & 92.3 & 96.9 & 98.6 & \textbf{99.1} \\
Passive Voice & 98.7 & 98.0 & 98.7 & 99.0 & 94.0 & 94.3 & \textbf{99.7} \\
\bottomrule
\end{tabular}
}
\end{table*}

Table~\ref{tab:trans} shows the per-transformation breakdown. 
Passive voice (surface control) remains the easiest transformation for most models, confirming that surface rephrasing is easy. 
Contrapositive and double negation, which add or reposition negation operators, cause the steepest drops for GPT-5.4-mini. 
The gap between the weakest logical transformation and passive voice is 26.3pp for GPT-5.4-mini but only 1.7pp for o4-mini, suggesting reasoning-optimized models handle operator-level rewrites more robustly.

Quantifier rewrite ($n{=}60$) is more variable, with the weakest results appearing for GPT-5.4-mini(r) (70.0\%) and GPT-5.4(r) (78.3\%), foreshadowing the category-level failures discussed below.

\begin{figure}[h]
    \centering
    \includegraphics[width=1.0\linewidth]{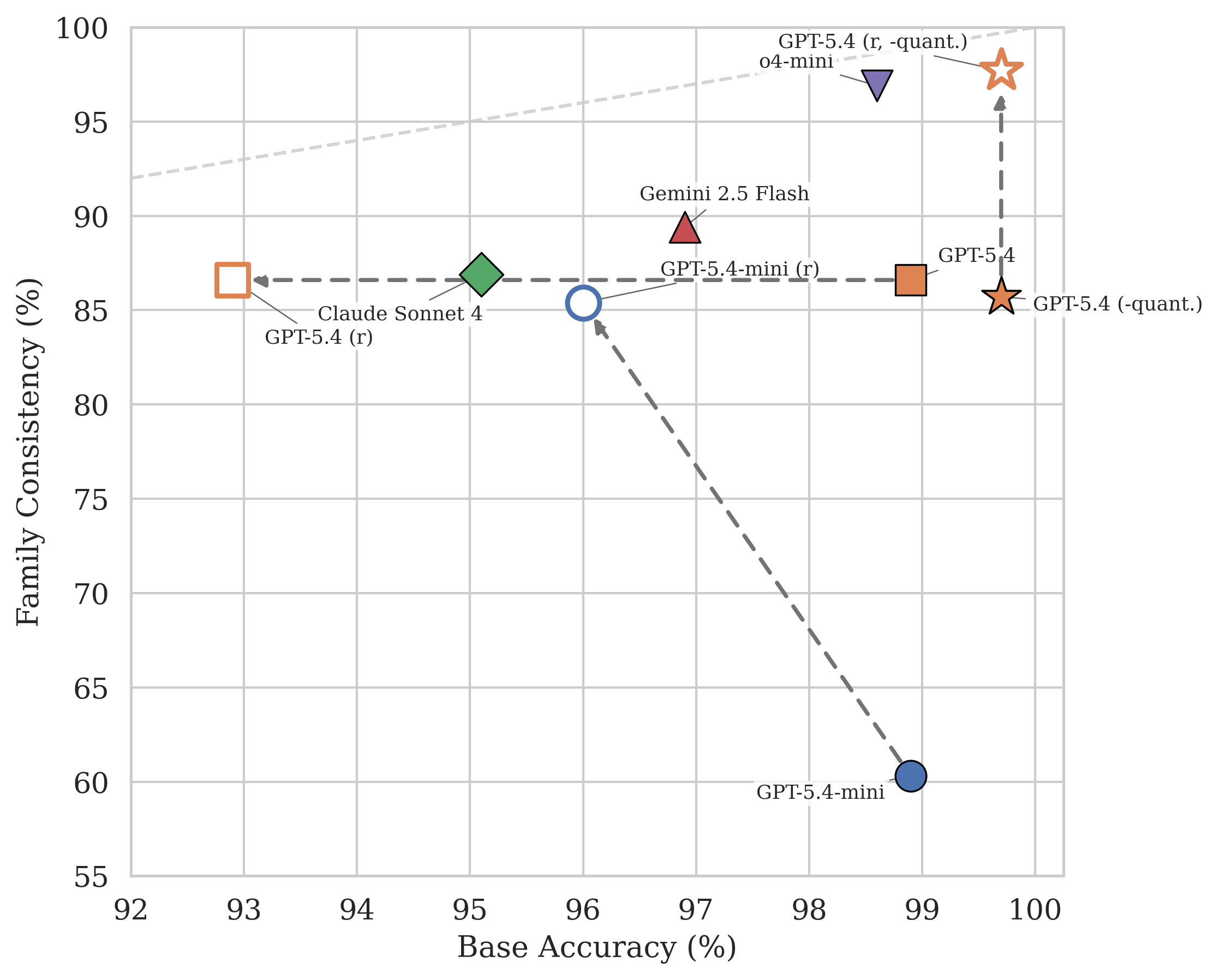}
    \caption{Accuracy vs. family consistency (zero-shot). Dashed arrows show the effect of enabling reasoning on GPT-5.4 and GPT-5.4-mini: 5.4-mini moves up and left, while 5.4 loses accuracy with no net gain in family consistency. Star markers show 5.4 and 5.4(r) after removing quantifier reasoning; without that category, 5.4(r) moves to 99.7\% accuracy and 97.7\% family consistency.}
    \label{fig:scatter}
\end{figure}

\subsection{Effect of Reasoning Ablation}

GPT 5.4 and 5.4-mini support a \texttt{reasoning\_effort} parameter (default: \texttt{none}). Enabling reasoning substantially improves GPT-5.4-mini as family consistency rises from 60.3\% to 85.4\% (+25.1pp), and double negation improves from 84.6\% to 92.3\%. However, base accuracy drops from 98.9\% to 96.0\%.
GPT-5.4, however, does not improve overall: family consistency remains 86.6\% while base accuracy drops from 98.9\% to 92.9\%.
Table~\ref{tab:gpt54_category} shows why. Reasoning improves propositional logic (86$\to$100\%), nested negation (50$\to$94\%), and conditional fallacies (84$\to$98\%), but collapses quantifier reasoning (92$\to$20\%), offsetting those gains on the category-balanced benchmark.
This suggests that reasoning effort improves logical invariance selectively rather than uniformly.
By contrast, o4-mini achieves both high accuracy (98.6\%) and high consistency (96.9\%), suggesting purpose-built reasoning models better avoid this tradeoff.
\begin{table}[h]
\centering
\small
\resizebox{\columnwidth}{!}{%
\begin{tabular}{lccc}
\toprule
\textbf{Configuration} & \textbf{Base Acc.} & \textbf{Family\ Cons.} & \textbf{Double.\ Neg.} \\
\midrule
GPT-5.4-mini (no reasoning) & 98.9\% & 60.3\% & 84.6\% \\
GPT-5.4-mini (reasoning=high) & 96.0\% & 85.4\% & 92.3\% \\
GPT-5.4 (no reasoning) & 98.9\% & 86.6\% & 93.1\% \\
GPT-5.4 (reasoning=high) & 92.9\% & 86.6\% & 95.4\% \\
o4-mini & 98.6\% & \textbf{96.9\%} & \textbf{98.0\%} \\
\bottomrule
\end{tabular}
}
\caption{Reasoning notably improves GPT-5.4-mini, but GPT-5.4's overall family consistency is unchanged because quantifier failures offset gains elsewhere.}
\label{tab:reasoning}
\end{table}

\begin{table}[h]
\centering
\small
\begin{tabular}{lcc}
\toprule
\textbf{Category} & \textbf{GPT-5.4} & \textbf{GPT-5.4 (r)} \\
\midrule
Propositional Logic & 86.0 & 100.0 \\
Syllogistic Reasoning & 94.0 & 94.0 \\
De Morgan & 100.0 & 100.0 \\
Quantifier Reasoning & 92.0 & 20.0 \\
Multi-hop Inference & 100.0 & 100.0 \\
Nested Negation & 50.0 & 94.0 \\
Conditional Fallacies & 84.0 & 98.0 \\
\bottomrule
\end{tabular}
\caption{Per-category family consistency (\%) for GPT-5.4 with and without reasoning on the 350-family benchmark. Reasoning sharply improves nested negation and conditional fallacies, but collapses quantifier reasoning, which explains why overall family consistency remains 86.6\% in both settings.}
\label{tab:gpt54_category}
\end{table}

\paragraph{Why quantifier reasoning collapses under high reasoning.} Inspection of the outputs suggests an overcautious hedging failure rather than systematic logical reversal. On the 50 quantifier families (250 questions), GPT-5.4 without reasoning answers ``Cannot be determined'' only 5 times, and all 5 are incorrect; these errors account for all 4 inconsistent families in that category. With \texttt{reasoning\_effort=high}, GPT-5.4 answers ``Cannot be determined'' 119 times, and all 119 are incorrect, dropping quantifier question-level accuracy from 98.0\% (245/250) to 52.4\% (131/250). All 40 quantifier families marked inconsistent under the family-consistency metric contain at least one such incorrect abstention. These abstentions cluster in double negation (38/50 questions), quantifier rewrite (26/50), negation flip (26/50), and even the base question itself (24/50), suggesting that higher reasoning effort often treats valid universal/existential inferences as undetermined. Consistent with this interpretation, Figure~\ref{fig:scatter} shows that when quantifier reasoning is removed, GPT-5.4(r) moves from (92.9, 86.6) to (99.7, 97.7) in accuracy/consistency space, while GPT-5.4 moves from (98.9, 86.6) to (99.7, 85.7).

\subsection{Prompting and Category Analysis}

\paragraph{CoT has minimal effect.} We ran GPT-5.4 and GPT-5.4-mini with chain-of-thought prompting on all 350 families. 
For family consistency, GPT-5.4-mini changes by +1.4pp (60.3$\to$61.7\%), GPT-5.4 changes by +0.0pp (86.6$\to$86.6\%), and GPT-5.4(r) changes by +0.0pp (86.6$\to$86.6\%); none are significant at $\alpha{=}0.05$, suggesting the consistency gap reflects model capabilities and not prompting techniques.

\paragraph{Category breakdown.} For GPT-5.4 zero-shot consistency, nested negation (50\%) is the hardest category, followed by conditional fallacies (84\%) and propositional logic (86\%). Quantifier reasoning reaches 92\%, syllogisms 94\%, and both De Morgan and multi-hop reach 100\%. Table~\ref{tab:gpt54_category} shows that reasoning mainly helps nested negation and conditional fallacies, but introduces a quantifier-specific failure. 

\section{Discussion and Conclusion}

\paragraph{Why negation is hard.} 
LLMs process negation compositionally but unreliably. 
Essentially, this means that even when single negation is handled well, stacking negations (``it is not the case that X is not Y'') leads to systematic failures. 
This matches with psycholinguistic findings on human negation processing \citep{kaup2007processing}, though the effect is substantially larger in LLMs.

\paragraph{Practical implications.} 
In applications requiring multi-query coherence (legal reasoning, medical diagnosis, knowledge bases), a model that contradicts itself across equivalent formulations cannot be trusted, even with high per-question accuracy. 
Consistency-aware evaluation should complement accuracy-based benchmarks.

\paragraph{Limitations.} Our benchmark is template-generated (350 families), and all families were manually audited and corrected for logical and grammatical errors. 
Although the benchmark is now larger and category-balanced, it still measures controlled reformulations rather than fully natural language variation. Our validation protocol is based on manual audit rather than a formal multi-annotator naturalness study, so claims about linguistic realism should be interpreted conservatively. 
We evaluate primarily zero-shot prompting with a single template.
In addition, our benchmark covers only binary and ternary-answer questions (Yes/No/Cannot be determined), which may not capture inconsistency in open-ended or multi-step reasoning settings where negation composition plays out over longer inference chains.

\paragraph{Conclusion.} 
In this paper, we show that high accuracy does not imply logical invariance. 
However, we believe that the most interesting finding is not just the accuracy-consistency gap itself, but where explicit reasoning helps and where it fails.
Enabling explicit reasoning closes much of this gap for GPT-5.4-mini (60.3$\to$85.4\%), but not for GPT-5.4 overall, where we observe that the gains on nested negation and conditional fallacies are canceled by a severe drop on quantifier reasoning.
Whether this reflects a different training signal that occurred due to Reinforcement Learning finetuning or something structural about how reasoning is integrated, such as surface-sensitive heuristics that break under operator-level reformulations, we believe this poses an important question to explore.
In the future, we aim to understand this mechanism and whether consistency-aware training can close the gap without losing accuracy.

\bibliography{references}
\bibliographystyle{plainnat}

\clearpage
\appendix
\onecolumn

\section{Dataset Details}

\begin{table}[h]
\centering
\small
\begin{tabular}{lcc}
\toprule
\textbf{Category} & \textbf{Families} & \textbf{Questions} \\
\midrule
Propositional Logic & 50 & 250 \\
Syllogistic Reasoning & 50 & 250 \\
De Morgan's Laws & 50 & 250 \\
Quantifier Reasoning & 50 & 250 \\
Multi-hop Inference & 50 & 250 \\
Nested Negation & 50 & 250 \\
Conditional Fallacies & 50 & 250 \\
\midrule
\textbf{Total} & \textbf{350} & \textbf{1750} \\
\bottomrule
\end{tabular}
\caption{Dataset statistics for \ours{}.}
\label{tab:dataset_stats}
\end{table}

\paragraph{Example family (propositional logic).}
\textbf{Base:} ``If it is raining, then the ground is wet. It is raining. Is the ground wet?'' (Yes).
\textbf{Contrapositive:} ``If the ground is not wet, then it is not raining. It is raining. Is the ground wet?'' (Yes).
\textbf{Double negation:} ``...Is it NOT the case that the ground is NOT wet?'' (Yes).
\textbf{Negation flip:} ``...Is the ground not wet?'' (No).
\textbf{Passive:} ``...Is the ground made wet?'' (Yes).

\paragraph{Example family (nested negation).}
\textbf{Base:} ``Is it true that cats are mammals?'' (Yes).
\textbf{Triple negation:} ``Is it NOT the case that it is NOT true that it is NOT the case that cats are mammals?'' (No --- $\neg\neg\neg P = \neg P$).
\textbf{Double negation:} ``Is it NOT the case that it is NOT true that cats are mammals?'' (Yes).
\textbf{Contrapositive:} ``If someone claimed cats are not mammals, would that claim be incorrect?'' (Yes).
\textbf{Passive:} ``Based on established facts, is the statement `cats are mammals' accurate?'' (Yes).

\paragraph{Per-category family consistency for GPT-5.4 (\%).}
For zero-shot GPT-5.4, propositional logic reaches 86.0, multi-hop 100.0, De Morgan 100.0, syllogistic reasoning 94.0, quantifier reasoning 92.0, nested negation 50.0, and conditional fallacies 84.0. The matched GPT-5.4 reasoning-ablation matrix is reported in Table~\ref{tab:gpt54_category}.

\paragraph{Construction and validation details.} Benchmark construction proceeded in two stages. We first created a base-only review set containing all 350 source questions, then manually reviewed that set for logical correctness, answer-label correctness, and obvious wording problems before generating any reformulations. After reformulations were added, we manually inspected families again to verify that each transformation either preserved or reversed the answer as intended, that the final wording remained grammatical, and that category-specific transformations were applied only where appropriate. This process served as internal quality control; we do not report separate inter-annotator agreement or a standalone human-naturalness evaluation.

\section{Experimental Details}
\label{app:details}

\paragraph{Model identifiers.} All API calls were made in April 2026 using the following model identifiers: \texttt{gpt-5.4}, \texttt{gpt-5.4-mini}, \texttt{claude-sonnet-4-20250514}, \texttt{gemini-2.5-flash}, \texttt{o4-mini}. Temperature is set to $T{=}0$ for all models except o4-mini, which does not support temperature control. Code, data, and all model outputs will be made public upon acceptance for reproducibility.

\paragraph{Prompts.} For zero-shot evaluation:

\begin{quote}
\small \texttt{You are a logical reasoning assistant. Answer the question with exactly one of: ``Yes'', ``No'', or ``Cannot be determined''. Give your final answer on a new line starting with ``Answer: ''.}
\end{quote}

\noindent For chain-of-thought evaluation:

\begin{quote}
\small \texttt{You are a logical reasoning assistant. Think step by step about the logical structure of the problem. Then give your final answer on a new line starting with ``Answer: '' followed by exactly one of: ``Yes'', ``No'', or ``Cannot be determined''.}
\end{quote}

\paragraph{Answer extraction.} Answers are extracted with a regex-first heuristic. We first search for explicit final-answer patterns such as \texttt{Answer: Yes}, \texttt{final answer is No}, bold-formatted answers, and conclusion phrases (``therefore,'' ``thus,'' ``in conclusion'') followed by one of the allowed labels. If those fail, we inspect the last three sentences for ``cannot be determined''-style phrases, then apply a lightweight fallback over Yes/No cues in the ending of the response and the first word. If no answer is found, the response is marked \texttt{UNCLEAR} and excluded from both accuracy and consistency calculations. Across the main runs, Gemini 2.5 Flash produced 5 such extraction failures.

\paragraph{Full zero-shot transformation breakdown.} Table~\ref{tab:transformation} reports the full zero-shot per-transformation results, including De Morgan and quantifier rewrite, which are omitted from the compact core-transformation table in the main text.

\begin{table*}[t]
    \centering
    \caption{Per-transformation consistency (\%) across the 350-family benchmark (zero-shot). Each cell shows the percentage of families where the model answered both the base question and its reformulation correctly for that transformation type. Best values in \textbf{bold}.}
    \label{tab:transformation}
    \footnotesize
    \setlength{\tabcolsep}{3pt}
    \begin{tabular*}{\textwidth}{@{\extracolsep{\fill}}lccccc@{}}
    \toprule
    \textbf{Transformation} & \textbf{GPT-5.4-mini} & \textbf{GPT-5.4} & \textbf{Claude Sonnet 4} & \textbf{Gemini 2.5 Flash} & \textbf{o4-mini} \\
    \midrule
        Contrapositive & 72.4 & 93.8 & 92.8 & 95.9 & \textbf{98.6} \\
        De Morgan & \textbf{100.0} & \textbf{100.0} & 96.0 & \textbf{100.0} & \textbf{100.0} \\
        Double Negation & 84.6 & 93.1 & 93.1 & 96.0 & \textbf{98.0} \\
        Negation Flip & 94.6 & 98.6 & 96.9 & 98.6 & \textbf{99.1} \\
        Passive Voice & 98.7 & 98.7 & 94.0 & 94.3 & \textbf{99.7} \\
        Quantifier Rewrite & 90.0 & 91.7 & \textbf{93.3} & 83.1 & 90.0 \\
    \midrule
        Overall & 60.3 & 86.6 & 86.9 & 89.4 & \textbf{96.9} \\
    \bottomrule
    \end{tabular*}
    \end{table*}

\section{Family Structure by Category}
\label{app:structure}

Table~\ref{tab:family_structure} specifies which transformations apply to each question category. Each family receives exactly 4 reformulations from its applicable set.

\begin{table}[t]
\centering
\footnotesize
\resizebox{\textwidth}{!}{%
\begin{tabular}{lcccccc@{}}
\toprule
\textbf{Category} & \textbf{Contrapositive} & \textbf{Double Negation} & \textbf{Negation Flip} & \textbf{Passive} & \textbf{DeMorgan.} & \textbf{Quantifier Reasoning} \\
\midrule
Propositional Logic & \checkmark & \checkmark & \checkmark & \checkmark & & \\
Syllogistic Reasoning & \checkmark & \checkmark & \checkmark & \checkmark & & \\
De Morgan's Laws & & \checkmark & \checkmark & \checkmark & \checkmark & \\
Quantifier Reasoning & \checkmark & \checkmark & \checkmark & & & \checkmark \\
Multi-hop Inference & \checkmark & \checkmark & \checkmark & \checkmark & & \\
Nested Negation & \checkmark & \checkmark & \checkmark & \checkmark & & \\
Conditional Fallacies & \checkmark & \checkmark & \checkmark & \checkmark & & \\
\bottomrule
\end{tabular}
}
\caption{Transformation types applied per question category.}
\label{tab:family_structure}
\end{table}

\section{Additional Example: Conditional Fallacy}
\label{app:fallacy_example}

\paragraph{Example family (affirming the consequent).}
\textbf{Base:} ``If an animal is a dog, then it is a mammal. This animal is a mammal. Is this animal necessarily a dog?'' (No; affirming the consequent fallacy).
\textbf{Contrapositive:} ``If something is not a mammal, then it is not a dog. This animal is a mammal. Is this animal necessarily a dog?'' (No).
\textbf{Double negation:} ``If an animal is a dog, then it is not the case that it is not a mammal. This animal is a mammal. Is this animal necessarily a dog?'' (No).
\textbf{Negation flip:} ``If an animal is a dog, then it is a mammal. This animal is a mammal. Can we rule out that this animal is a dog?'' (No. Cannot rule it out either).
\textbf{Passive:} ``Being a dog is sufficient for being a mammal. Given that this animal is a mammal, must it be a dog?'' (No).

\end{document}